\documentclass[letterpaper, 10 pt, conference]{ieeeconf}  %

\IEEEoverridecommandlockouts                              %

\usepackage{adjustbox}
\usepackage[linesnumbered,ruled,vlined]{algorithm2e}%
\usepackage{amsmath}
\usepackage{amssymb}
\usepackage{array}  %
\usepackage{authblk}
\usepackage[accsupp]{axessibility}  %
\usepackage{booktabs}
\usepackage{capt-of}
\makeatletter
\let\NAT@parse\undefined
\makeatother
\usepackage{cite}
\usepackage[font={small}]{caption}
\usepackage{colortbl}
\usepackage{epsfig}
\usepackage{inconsolata}  %
\usepackage{graphicx}
\usepackage{listings}
\usepackage{multicol}
\usepackage{multirow}
\usepackage{paralist}
\usepackage{pifont}
\usepackage{times}
\usepackage{tabularx}
\usepackage[table]{xcolor}
\usepackage{svg}

\definecolor{flodarkpurple}{rgb}{0.288,0.1196,0.7}

\definecolor{amber}{rgb}{1.0, 0.75, 0.0}

\usepackage[backref=page,breaklinks,colorlinks,bookmarks,citecolor=flodarkpurple]{hyperref}

\usepackage{booktabs, multirow} %
\usepackage{soul}%
\usepackage{changepage}
\usepackage{threeparttable}
\usepackage{caption}
\usepackage{subcaption}

\newcommand{\coolname}{\textit{ALT-Pilot}}
\newcommand{\webpage}{https://navigate-anywhere.github.io/ALT-Pilot/}

\newcommand{\authorhref}[3][flodarkpurple]{\href{#2}{\color{#1}{#3}}}

\title{\Large \bf 
{ALT}-Pilot: \textbf{A}utonomous navigation with \textbf{L}anguage augmented \textbf{T}opometric maps \\
\vspace{0.30em}
\large{\href{\webpage}{\color{violet}{\texttt{\webpage}}}}
}

\author{
\authorhref{webpage}{Mohammad Omama}$^{1}$, 
\authorhref{webpage}{Pranav Inani}$^{*2}$, 
\authorhref{webpage}{Pranjal Paul}$^{*2}$, 
\authorhref{webpage}{Sarat Chandra Yellapragada}$^{2}$,
\\
\authorhref{webpage}{Krishna Murthy Jatavallabhula}$^{\dagger3}$, 
\authorhref{webpage}{Sandeep Chinchali}$^{\dagger1}$, and 
\authorhref{webpage}{Madhava Krishna}$^{\dagger2}$
\\
$^{1}$\href{https://www.utexas.edu/}{The University of Texas at Austin}, 
$^{2}$\href{https://robotics.iiit.ac.in/}{International Institute of Information Technology, Hyderabad}, \\
$^{3}$\href{https://web.mit.edu}{Massachusetts Institute of Technology}
\thanks{*Co-second authors $\dagger$Equal advising }%
}

\begin{document}

\makeatletter
\let\@oldmaketitle\@maketitle
\renewcommand{\@maketitle}{\@oldmaketitle
\centering
\includegraphics[width=0.96\linewidth]{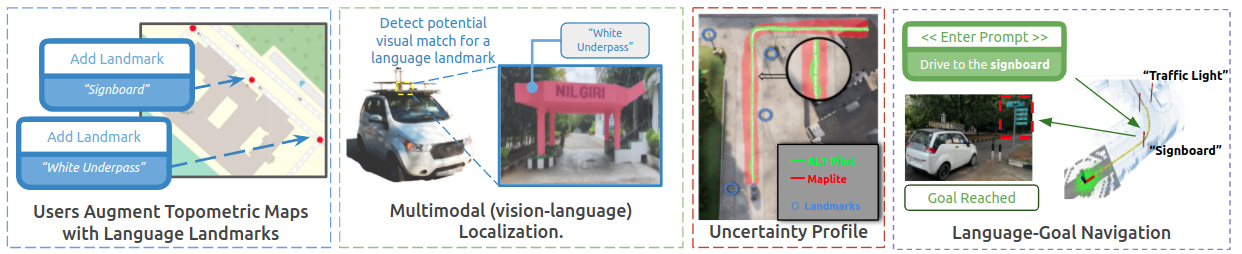}
\captionof{figure}{
\textbf{\coolname{}} enables navigating to any point on a road network specified via a GPS waypoint, map click, or language prompt without assuming a HD map of the environment, relying instead on crowdsourced topometric maps like OpenStreetMap (OSM).
\textit{Left to right}: We allow users to augment these topometric maps by clicking anywhere on the map and adding a \emph{language-only landmark} -- a brief text tag describing a distinct static object in the scene, such as a street sign or a roundabout.
At the core of \coolname{} is a probabilistic multimodal localization algorithm that bridges LiDAR, image, and language to estimate a posterior distribution of vehicle poses in the map frame (\textit{top-right}).
This allows us to account for erroneously specified landmarks, and provides a global localization estimate useful for route (global) planning, which we couple with onboard sensors and perception for trajectory planning and control.
From the uncertainty profile plot, notice how \coolname{} is more \emph{certain} over the entire trajectory; compared to Maplite~\cite{maplite}, a state-of-the-art topometric navigation approach for AD.
Additionally, the language-augmented maps enable navigating to open-vocabulary goals specified as text queries.
}
\label{fig:splash}
\vspace{-0.4cm}
}
\makeatother

\maketitle
\thispagestyle{empty}
\pagestyle{empty}

\begin{abstract}

We present an autonomous navigation system that operates without assuming HD LiDAR maps of the environment.
Our system, \coolname{}, relies only on publicly available road network information and a sparse (and noisy) set of crowdsourced language landmarks.
With the help of onboard sensors and a language-augmented topometric map, \coolname{} autonomously pilots the vehicle to any destination on the road network.
We achieve this by leveraging vision-language models pre-trained on web-scale data to identify potential landmarks in a scene, incorporating vision-language features into the recursive Bayesian state estimation stack to generate global (route) plans, and a reactive trajectory planner and controller operating in the vehicle frame.
We implement and evaluate \coolname{} in simulation and on a real, full-scale autonomous vehicle and report improvements over state-of-the-art topometric navigation systems by a factor of 3$\times$ on localization accuracy and 5$\times$ on goal reachability.

\end{abstract}

\setcounter{figure}{1} %

\section{Introduction}
\label{sec:intro}

Existing industry-led autonomous driving (AD) systems rely crucially on high-definition (HD) maps, meticulously constructed with centimeter-level precision.
These HD maps impose large compute and memory requirements for map creation and maintenance~\cite{how-tokeep-hd-maps-upto-date}, which has insofar limited the applicability of such AD systems to urban areas
where HD map generation and maintenance are both computationally and economically justifiable. Moreover, the inherent memory demands and human resource investments for HD map creation severely limit the scalability of these systems.

\subsection{Related Work}

\textbf{Autonomous driving without HD maps.} To eliminate the dependence on HD maps, existing approaches~\cite{maplite-precursor, maplite, sensors2022lidar-osm-pf, ninan2023road} use crowdsourced \emph{topometric} maps like OpenStreetMap (OSM) -- topologically accurate, but metrically imprecise road networks.
These approaches employ topometric maps for global route planning and integrate them with onboard sensors for state estimation, trajectory planning, and control.
While these systems have enabled navigation to destinations near road intersections, their localization accuracy deteriorates as the vehicle ventures farther away (see Fig.~\ref{fig:splash}).

\textbf{Vision-language models.} Recent advances in vision-language models~\cite{radford2021clip,lseg,openseg,dong2023maskclip} have demonstrated promising capabilities for both 2D and 3D perception~\cite{conceptfusion, huang23vlmaps,clip-fields,lerf}.
These approaches enable navigation to open-vocabulary goals, usually specified in the form of text prompts, by aligning semantic concepts across language and images.
However, these approaches too, require significantly large memory and/or compute footprints, owing to the storage of high-dimensional neural embeddings~\cite{conceptfusion,lseg} or require a large number of queries to a neural field~\cite{clip-fields,lerf}; both impractical for AD.

\subsection{Summary of Our Contributions}

In this paper, we address two primary questions:
\begin{enumerate}
    \item What kinds of map representations enable autonomous navigation to any destination on a road network\footnote{Current best approach~\cite{maplite} only enables reliable navigation to intersections, but not arbitrary destinations along a road.}, without relying on HD lidar maps?
    \item How do we accurately localize on this map representation, while accounting for errors arising due to the map representation, and due to imperfect perception and control?
\end{enumerate}

\begin{figure*}
     \centering
      \includegraphics[width=0.95\textwidth]{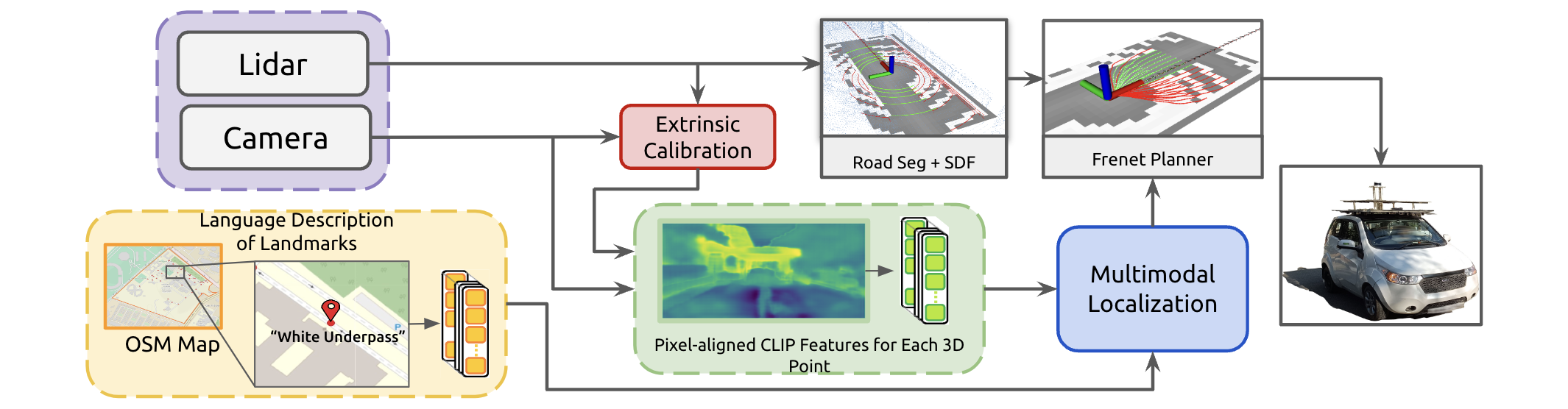}
     \caption{\textbf{Pipeline}: \coolname{} uses OSM maps augmented with language descriptions of landmarks. We precompute the CLIP (language) descriptors for each of these landmarks. At inference time, the LiDAR and Camera information is used to generate per-pixel CLIP (image) descriptors used in our multimodal localization algorithm (\ref{sec:multimodal-loc}). The LiDAR data is used to segment points on the road and generate a costmap (\ref{sec:lidar-percp}). The localization and perception information then goes to a planning module which generates actuation commands. }
     \label{fig:pipeline}
     \vspace{-1em}
\end{figure*}

To address the former question, we propose to use \textbf{language-augmented topometric maps} -- where we augment off-the-shelf OSM road networks with a small number of language landmarks (such as street signs, traffic signals, and other persistent features in the environment useful for localization).
The landmarks are \emph{language-only}, meaning we store only a language tag; but no other information (such as appearance or size information) is used.
This makes the map extremely convenient to build, mirroring OSM's annotation methodology, where users simply click on a displayed road network and enter a brief language description\footnote{We demonstrate robustness to errors in landmark location, false positives/negatives, incorrect language descriptions in Sec.~\ref{sec:results}.}.

Towards the latter, we propose \textbf{\coolname{}}, an autonomous navigation system capable of piloting a vehicle to any destination over a language-augmented topometric map.
We develop a multimodal localization pipeline that efficiently bridges multiple modalities such as LiDAR (for road perception), RGB stream (for vision-language feature extraction), and language (for landmark description) to safely and accurately navigate to any location on the road network.
Besides enabling localization and autonomous navigation, language landmarks also enable open-vocabulary navigation based on goals described in language.

To summarize, our \textbf{key contributions} are:
\begin{enumerate}
    \item \coolname{}: An autonomous navigation system that is completely devoid of HD LiDAR maps, relying instead on language-augmented topometric maps; enabling accurate localization and open-vocabulary navigation.
    \item A probabilistic multimodal localization system bridging vision, lidar, and language to estimate a posterior over vehicle poses.
\end{enumerate}

Our \textbf{key results} are as follows:
\begin{enumerate}
    \item \coolname{} localizes to a much higher degree of accuracy, \emph{and} over longer stretches of the road network, compared to state-of-the-art topometric localization approaches such as MapLite~\cite{maplite}.
    \item Based on evaluations on a \textbf{real-world}, full-size autonomous car over 3 km of driving data, and more elaborate evaluations on the CARLA~\cite{dosovitskiy2017carla} simulator, we outperform current art by a factor of 3$\times$ on localization precision and 5$\times$ on goal reachability.
\end{enumerate}

\section{Multimodal Topometric Maps}
\label{sec:multimodal-topometric-maps}

This work focuses on autonomous driving platforms equipped with a lidar sensor and extrinsically calibrated color cameras.
To enable our system to be as general as possible, our multimodal topometric map is minimalist by design and is only intended to be used as a \emph{guide} to aid global planning and localization.
Our map comprises two key elements: the road network, and (approximately) geotagged language landmarks.

The road network is a graph where the presence of an edge between two nodes $N_A$ and $N_B$ indicates that there exists a navigable path between the two nodes.
These road networks are already openly available for the entire world. For instance, using the popular OpenStreetMap API, the entire road network for the continental United States may be downloaded and stored in about 1 GB of memory.
While prior approaches have investigated using solely the OSM's road network\cite{maplite,sensors2022lidar-osm-pf}, such approaches are only suited to navigate to intersections, owing to their localization algorithm only being able to precisely localize at intersections. 

In all of the real-world experiments in this work, we preprocess and extract only the road network from the OpenStreetMap (OSM). We augment the road network with approximately geotagged point \emph{landmarks}. Each point landmark has a geolocation that places it relative to the UTM coordinate frame for the local UTM zone; and a short (1-3 word) language tag that briefly describes the landmark. Ideal choices for such landmarks are traffic signs, roundabouts, underpasses, benches, poles, or other stationary objects in the environment. Using crowdsourced maps such as OSM enables these landmarks to be created by any user via their web interface; resulting in desirable scaling properties and minimal human intervention.

\section{Approach}
\label{sec:approach}

Using the aforementioned multimodal topometric map representation, we design an autonomous navigation system (see Fig-\ref{fig:pipeline}) that can safely navigate to any location on the road network.
The system accepts a geolocation of the destination as input, and navigates toward the goal with the help of the following subsystems:
\begin{enumerate}
    \item the \textbf{LiDAR Perception} subsystem consumes a laser scan, segments the road plane, and constructs a Euclidean Signed Distance Field (ESDF) for use in trajectory planning and control.
    \item our novel \textbf{Multimodal Localization} subsystem extracts cues from the LiDAR and Image data to localize against the road network and text landmarks encoded in the topometric map.
    \item our \textbf{Path Planning}, \textbf{Trajectory Planning}, and \textbf{Controllers} are based on prior work \cite{maplite, frenet-planner} , with minor modifications necessary for integration with the other subsystems.
\end{enumerate}

\subsection{LiDAR Perception}\label{sec:lidar-percp}

Each input lidar scan is first segmented using the segmentation method proposed in~\cite{roadfilt2022horv} to extract all points on the road plane.
We then construct a (binary) occupancy grid in a bird's-eye view, by extracting a $30 \ m \times 15 \ m$ region centered around the vehicle. This is used to compute a distance field $\mathcal{D}: \mathbb{R}^2 \mapsto \mathbb{R}^+$, where for each grid location $\mathit{x} \in \mathbb{R}^2$, $\mathcal{D}(\mathit{x})$ denotes the distance from $\mathit{x}$ to the closest obstacle.
For use in our localization and variational trajectory planner, we compute a Euclidean signed distance field $\mathcal{E}: \mathbb{R}^2 \mapsto \mathbb{R}$, where $\mathcal{E}(\mathit{x}) = \mathcal{D}(\mathit{x}) - \mathcal{D}(1 - \mathit{x})$ is the difference between the distance field and the \emph{inverse} distance field. The ESDF $\mathcal{E}$ forms the basis for our probabilistic multimodal localization subsystem.

\subsection{Multimodal Localization across LiDAR, Image, and Text} \label{sec:multimodal-loc}

Accurately localizing the vehicle within the topometric map is critical to the success of autonomous navigation.
In our context, the localization subsystem must bridge the gap across multiple modalities: the topometric map (a graph augmented with landmarks and language tags), and onboard sensor data (lidar, camera, IMU, and other odometry sources).
We develop a probabilistic multimodal localization approach based on particle filters. For a refresher on particle filters, we refer the interested reader to \cite{thrun2002probabilistic, fox1999monte, fox2003adapting}.

We want to estimate the posterior distribution, at time step $t$, over the vehicle state $\mathbf{x_t} = (x_t, y_t, \theta_t) \in SE(2)$.
This posterior is a set of $N$ hypotheses (particles) $\mathcal{X}_t = \mathbf{x}^{[1]}_t, \mathbf{x}^{[2]}_t, \cdots, \mathbf{x}^{[N]}_t$, initialized uniformly and randomly over the entire map. At each time step, given the control $u_t$, the set of particles is updated by resampling according to a motion model $p$, derived from either the exteroceptive (lidar, camera) or the proprioceptive (IMU, wheel encoders) sensors.
$$
\mathbf{x}^{[m]}_t \sim p(\mathbf{x}_t^{[m]} | u_t, \mathbf{x}^{[m]}_{t-1})
$$
In our experiments, motion model updates are performed solely from off-the-shelf lidar odometry~\cite{shan2018lego}.
Sampling solely based on the motion model (dead reckoning) results in the state estimation error increasing unboundedly over time.

To mitigate this, we define an observation likelihood that considers all sensing modalities and map context.
Recall that CLIP features are aligned across vision and lanugage, and may be compared across domains by means of a cosine similarity measure. We precompute CLIP text features for all landmarks in the map to create a set $K$ where each $k_i \in K$ has an associated position $k_i^{pos}$ and a language feature $k_i^{feat_l}$.

At run time, we first process the incoming RGB image using the pixel-aligned feature extraction approach proposed in ConceptFusion~\cite{conceptfusion} to extract dense CLIP features for each pixel in the image. For each particle denoted as $\mathbf{x}_t^{[m]}$, we employ its pose hypothesis to extract a subset of landmarks $\mathrm{\tilde{K}} \subseteq K$ from the topometric map that may be visible in the image. Given our extrinsic calibration of the LiDAR-Camera setup, we can determine the collection of LiDAR points that project to valid image pixels. We create a set $L$ of such lidar points (projected to map frame using the particle hypothesis) such that each $l_i \in L$ has a vision feature $l_i^{feat_v}$ and a position $l_i^{pos}$ associated with it. We project the ESDF computed by the perception system to the map frame to get $\tilde{\mathcal{E}}$. Also, we create a set $R = (r_1, r_2, ..., r_v)$ of $v$ closest road points to the particle hypothesis.

Then, for each particle $\mathbf{x}_t^{[m]}$, we calculate an importance factor denoted as $w_t^{[m]}$ in accordance with Algorithm \ref{alg:obsmodel}. This importance factor is subsequently used in the particle resampling process. The algorithm has the following key components:

\begin{algorithm}
\caption{Calculation of the importance factor for a single particle hypothesis} \label{alg:obsmodel}
\KwData{Landmarks $ \mathrm{\tilde{K}} $, LiDAR points $ L $, ESDF in map frame $\mathcal{\tilde{E}}$, Road Points $R$}
\KwResult{Hypothesis Importance Factor $w^{[m]}_t$}
\SetKwFunction{ComputeCosineSim}{CosineSim}
Landmark Weight $w_{lm} \gets 0$ \

\ForEach{landmark $ \tilde{k}_j $ in $ \mathrm{\tilde{K}} $}{
    $ \text{Closest LiDAR Match} \ \tilde{l} \leftarrow $ null \
    
    $ \text{Maximum Cosine Similarity c} \leftarrow 0 $ \
    
    \ForEach{lidarPoint $ l_i $ in $ L $}{  \label{alg:line-sim-start}  
    
        $ \text{sim} \leftarrow $ \ComputeCosineSim{$ k_j^{feat_l} $, $ l_i^{feat_v} $} \
        
        \If{$ \text{sim} > \text{c} $}{
        
            $ \tilde{l} \leftarrow l_i $ \
            
            $ \text{c} \leftarrow \text{sim} $ \ \label{alg:line-sim-end} 
            
        }
    }
    Distance Reciprocal $\tilde{d} \gets \dfrac{1}{ ||\tilde{l}_j^{pos}-\tilde{k}_j^{pos}|| + \epsilon }$  \ \label{alg:line-dis-1}
    
    Distance Factor $d \gets \alpha + \dfrac{1}{ 1 + e^{-\beta\tilde{d}} }$ \ \label{alg:line-dis-2}
    
    $w_{lm} := w_{lm} + c \times d$ \    \label{alg:line-wlm}

}
Yaw Weight $w_{yaw} \gets  \sum_{j} \mathcal{\tilde{E}}(r_j)$  \  \label{alg:line-wyaw}

$w^{[m]}_t = w_{lm} + \lambda w_{yaw}$ \  \label{alg:line-final}
\end{algorithm}

\begin{itemize}
    \item For each landmark we calculate the cosine similarity value $c \in [0,1]$ of the closest matching lidar point $\tilde{l}$ (line \ref{alg:line-sim-start}-\ref{alg:line-sim-end}). It ensures that the observation model corrects itself only when pixels corresponding to the appropriate landmark description are observed, thus mitigating the impact of false-positive landmarks.
    \item In line \ref{alg:line-dis-1}-\ref{alg:line-dis-2}, we calculate the distance factor $d \in [0,1]$. We use generalized logistic functions \cite{richards1959flexible} with tunable parameters $\alpha$ and $\beta$ to control the temperature of the distance term and ensure that it's in the range [0,1]. The purpose of $d$ is to activate the observation model only when the landmark is actually located at that position, thereby reducing the influence of false-positive CLIP matches. 
    \item The terms $c$ and $d$ are multiplied and summed for all landmarks on line \ref{alg:line-wlm} to get the landmarks weight $w_{lm}$.
    \item On line \ref{alg:line-wyaw} we compare the ESDF projected on the map frame, $\tilde{\mathcal{E}}$, and the road geometry expected around that location to calculate how well the particle aligns with the road. If the yaw of the particle hypothesis is close to the ground truth yaw, then the road geometry $R$ and the orientation of the road observed in $\tilde{\mathcal{E}}$ will align leading to a larger value of $w_{yaw}$ and vice-versa.
    \item We sum $w_{lm}$ and $w_{yaw}$ with a tunable parameter $\lambda$ to get the importance factor  $w_t^{[m]}$ of the particle $\mathbf{x}_t^{[m]}$.
\end{itemize}
Sparse landmarks often prove insufficient for effective yaw correction, necessitating the incorporation of an additional ESDF component. The impact of $w_{yaw}$ is presented in section \ref{sec:locresults} while the ablation on false positives and negatives is studied in section \ref{sec:fpandfn}.

\subsection{Planning and Control}\label{sec-planningnC}
Given a vehicle's current pose estimate and a goal point in the topological map, we employ the A* algorithm \cite{a-star} to determine the shortest path and extract waypoints for the next $15 m$ stretch of road. To account for uncertainty in the map and our state estimates, we employ a Frenet planner~\cite{frenet-planner} that samples trajectories given the waypoints and the precomputed ESDF from the perception stage.
Each trajectory is evaluated based on proximity to waypoints and ESDF costs, ensuring collision-free and accurate tracking.
Concurrently, a Stanley controller~\cite{stanley} actively guides the vehicle along the planned trajectory, while a PID controller manages dynamic speed adjustments.

\section{Experiments, Results, and Analyses}\label{sec:results}

In this section, we evaluate \coolname{} across simulated and real-world topomeric navigation scenarios with respect to current art~\cite{maplite}.
Our experiments span topological, topometric, and metric localization, while also analyzing convergence times and planner reachability. Specifically, our experiments aim to address these primary questions:

\begin{enumerate}
    \item \textit{How accurately does \coolname{} localize over a topometric map?} This is measured by means of three metrics:
    \begin{enumerate}
        \item \textbf{Recall@K}: Jaccard similarity of the K nearest landmarks of the ground truth position and the estimated position. Measures topological localization accuracy.
        \item \textbf{Distance to Closest Landmark Region (DCLR)}: Measures the proximity of the estimated position to a landmark region, defined by a radius $r$ from the nearest actual landmark. If within $r$, the metric is zero; otherwise, it measures the distance to the region. This combines topological and metric aspects for a nuanced evaluation. Mathematically, it can be written as: $max(0, \sqrt{(p_x-l_x)^2 + (p_y-l_y)^2} - r)$,
        where $p_x$ and $p_y $ represent the position estimate, $l_x, l_y$ are the landmark coordinates, and $r$ is the radius of the landmark region. 
        \item \textbf{Absolute Position Error (APE)}: It's the Euclidean distance between the estimated and actual positions. Measures metric accuracy.
    \end{enumerate}
    \item \textit{How quickly does the vehicle achieve convergence in a global location setting?} To quantify this, we measure the distance traversed until our localization system (multimodal particle filter) has converged.
    \item \textit{How effectively does \coolname{} reach its intended destination in a closed-loop setting?} To assess this, we measure the distance from the vehicle's final location (where it declares task success/failure) to the true destiation.
\end{enumerate}
Additionally, we present qualitative results and case studies underscoring the benefits of augmenting OpenStreetMap (OSM) with language landmarks for language-guided navigation. We also investigate the robustness of our approach to noise in landmark locations, and false postives/negatives.

\begin{table}[!htp]\centering
\caption{\textbf{Absolute Position Error in Map Frame}. Comparison of various baselines on multiple maps with varying levels of map noise. ALT-Pilot(L) is a subset of our system (no yaw correction), while ALT-Pilot, our full system, achieves superior performance.}\label{tab:ape }
\scriptsize
\begin{tabular}{lrrrrrr}\toprule
\textbf{Scene} &\textbf{ Noise} &\textbf{DeadReck} &\textbf{MapLite} &\textbf{ALT-Pilot(L)} &\textbf{ALT-Pilot} \\\midrule
\textbf{CARLA1} &\textbf{0\%} &21.1 &10.3 &7.2 &\textbf{3.98} \\
\textbf{} &\textbf{10\%} &32.7 &15.5 &17.3 &\textbf{4.94} \\
\textbf{} &\textbf{15\%} &45.3 &24.7 &29.7 &\textbf{5.61} \\
\textbf{} &\textbf{20\%} &50.4 &24.4 &39.2 &\textbf{7.52} \\
\textbf{CARLA2} &\textbf{0\%} &10.2 &6.4 &6.8 &\textbf{2.3} \\
\textbf{} &\textbf{10\%} &13.8 &10.1 &7.4 &\textbf{3.1} \\
\textbf{} &\textbf{15\%} &17.3 &11.7 &6.6 &\textbf{3.7} \\
\textbf{} &\textbf{20\%} &20.4 &11.9 &7.1 &\textbf{5.1} \\
\textbf{CARLA3} &\textbf{0\%} &10.2 &7.4 &\textbf{4.1} &4.5 \\
\textbf{} &\textbf{10\%} &15.4 &11.3 &\textbf{6.5} &7.1 \\
\textbf{} &\textbf{15\%} &24.7 &15.3 &\textbf{7.2} &7.7 \\
\textbf{} &\textbf{20\%} &30.1 &18.4 &\textbf{8.9} &9 \\
\textbf{CARLA4} &\textbf{0\%} &23.2 &15.1 &9.8 &\textbf{3.2} \\
\textbf{} &\textbf{10\%} &35.2 &19.1 &14.1 &\textbf{6.65} \\
\textbf{} &\textbf{15\%} &48.3 &33.3 &20.7 &\textbf{6.17} \\
\textbf{} &\textbf{20\%} &66.9 &40.8 &30.5 &\textbf{8.22} \\
\textbf{} &\textbf{} & & & & \\
\textbf{Real1} &\textbf{0\%} &13.03 &9.94 &4.01 &\textbf{3.29} \\
\textbf{Real2} &\textbf{0\%} &16.38 &12.66 &3.89 &\textbf{3.88} \\
\bottomrule
\end{tabular}
\end{table}

\subsection{Experimental Setup and Baselines}
We assess localization performance across four CARLA scenarios (spread across three distinct town maps), averaging results across three start and end points. 
Given our reliance on crowd-sourced OpenStreetMap (OSM) data, which lacks HD map precision, to assess our system's resilience to map inaccuracies, we scale the road network by upto 20\% of its original size (to induce errors in road segment lengths, and exacerbate errors due to incorrectly positioned landmarks).
We also evaluate real-world performance on two maps collected at the IIIT Hyderabad campus, over trajectories spanning 3 km with several intersections and turns.

We compare the following approaches:
\begin{enumerate}
    \item \textbf{Dead Reckoning}: odometry is integrated over time (error grows unboundedly due to the lack of feedback)
    \item \textbf{MapLite}: Using the observation likelihood outlined in Maplite\cite{maplite} (the current-best approach to topometric localization), and extending it to a particle filter setting for fair evaluation.
    \item \textbf{\coolname\textit{(L)}}: Utilizing only language landmarks, without the yaw correction term (i.e., no ESDF)
    \item \textbf{\coolname}: Using our full system (language landmarks and ESDF) outlined in Algorithm 1
\end{enumerate}
We evaluate closed-loop performance (i.e., the system navigating in conjunction with state estimation, planning, and control) over two CARLA scenarios.

For the data collection of \textbf{Real1} and \textbf{Real2} maps, and for the closed-loop language-guided navigation experiments in Sec.~\ref{sec:langnavreal}, we employ our custom, fullsize drive-by-wire electric vehicle, outfitted with an extrinsically calibrated 16-channel Velodyne LiDAR and a RGB camera.

\begin{figure}[!htb]
    \centering
    \includegraphics[width=0.45\textwidth]{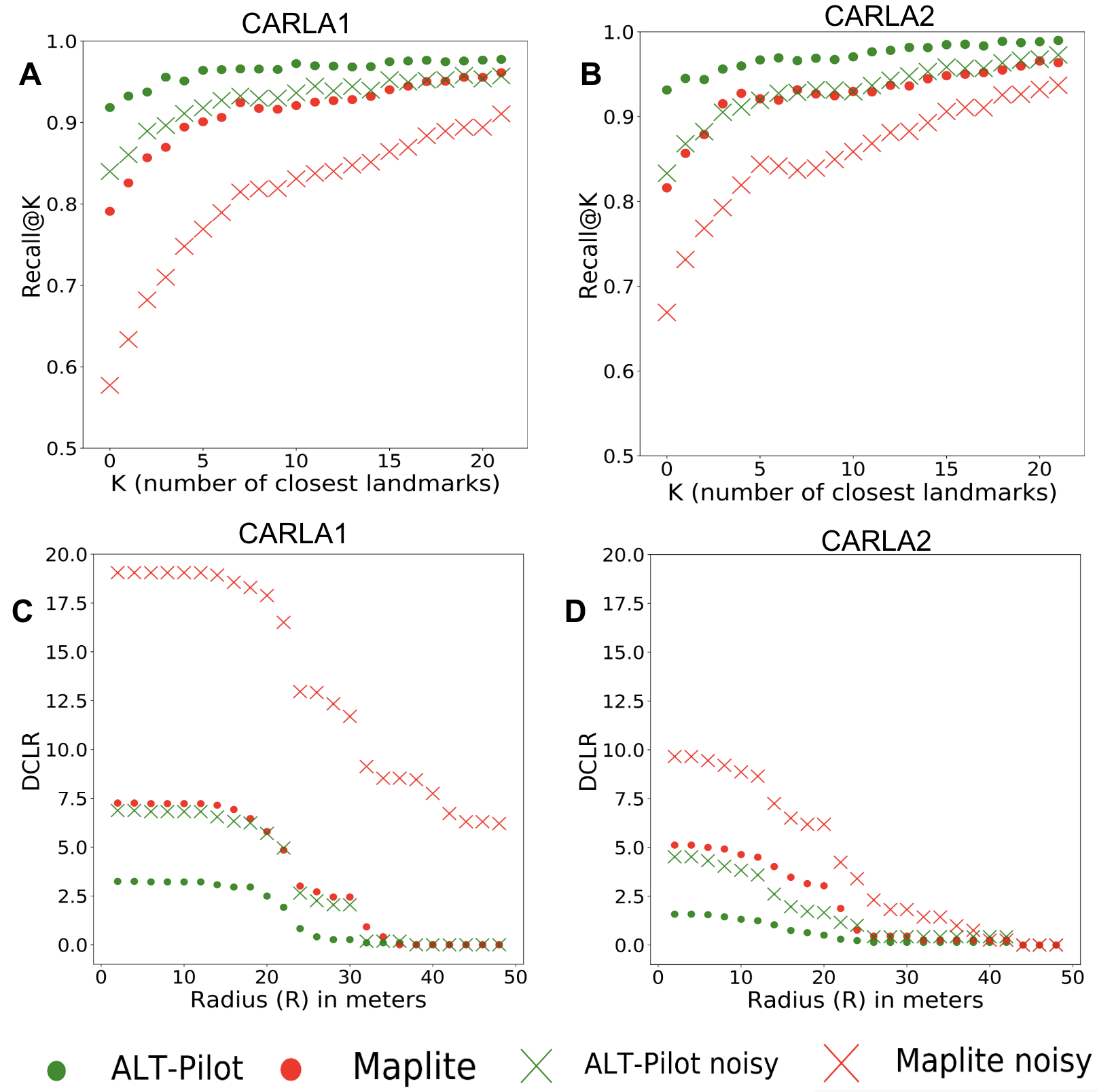}
    \caption{\textbf{Topological and Topometric localization on two CARLA maps.} \coolname \ outperforms MapLite in both topological and topometric localization. Further, it's more robust to noise in the map while MapLite is adversely affected by it. The impact of noise is more noticeable in the larger \textbf{CARLA1} map (panels A, C).}
    \label{fig:topo-logy-metric}
\end{figure}

\subsection{Localization Results}\label{sec:locresults}

\begin{figure}[!htb]
    \centering
    \includegraphics[width=0.45\textwidth]{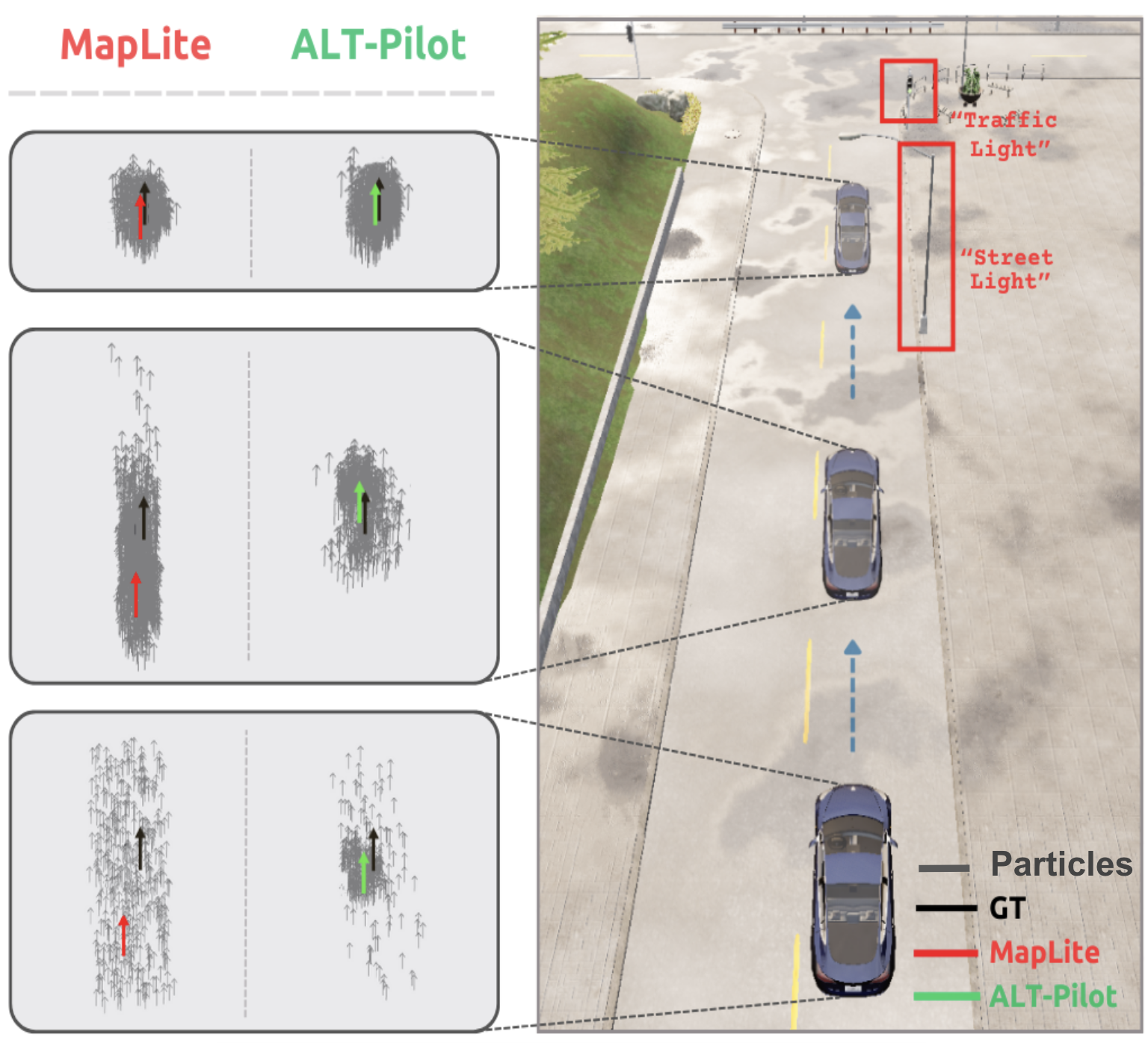}
    \caption{\textbf{Particle Filter Localization.} \coolname \ utilizes language descriptions of landmarks (red bounding boxes) for state estimation while MapLite only uses intersection information. \coolname \ is more accurate and certain throughout the run. MapLite shows comparable performance only near the intersection (top row).}
    \label{fig:qual}
\vspace{-5mm}
\end{figure}

In Fig-\ref{fig:qual}, we show a qualitative analysis of our approach against MapLite. \coolname \ utilizes language landmarks along the way, resulting in better state estimation while MapLite, which only uses intersection information, incurs substantial drift. Further \coolname \ has lower uncertainty throughout the run while MapLite only achieves low uncertainty near an intersection.

Table~\ref{tab:ape } compares the APE of our method, \coolname, with other baselines. \coolname \ consistently excels in APE. In large maps like `CARLA1' and `CARLA4', yaw correction notably boosts localization -- without the yaw correction, landmark sparsity in these maps will have resulted in poor localization performance (as can be seen in \coolname\textit{L}).

We further evaluate our performance by examining the topological performance using the `Recall@K' metric and the topometric performance with the `DCLR' metric. In Fig-\ref{fig:topo-logy-metric}[A, B] and Fig-\ref{fig:topo-logy-metric}[C, D], we present the results for `Recall@K' and `DCLR' on two CARLA maps, respectively. The metrics are important for downstream reachability tasks. Notably, as the size of the topological region increases, whether by enlarging K or R, both \coolname \ and MapLite displays comparable results. However, in smaller topological regions, \coolname \ outperforms, showcasing its relevance in addressing fine-grained reachability challenges.

Further \coolname \ is quite robust to map noise. By inducing 20\% map noise, Recall@K scores for both \coolname and MapLite drop by 0.1 and 0.23 in CARLA1 (\ref{fig:topo-logy-metric}[A]) and by 0.12 and 0.17 CARLA2 (\ref{fig:topo-logy-metric}[B]) respectively.
Similarly, the DCLR values increase by 3.8m and 11m in CARLA1 (\ref{fig:topo-logy-metric}[C]) and by 3m and 5m in CARLA2 (\ref{fig:topo-logy-metric}[D]) respectively.
The impact of noise is more observable in larger maps. 
Notice that \coolname{} degrades more gracefully compared to MapLite, which incurs more significant errors.

\subsection{Global Localization: Distance traveled until Convergence}
In Table \ref{tab:convergence}, we present the distance traveled by the vehicle until the particle filter \emph{converges}. The particle filter is deemed converged when the median of all particles is within 5m ground-truth location, and their variance is less than 10m. In all these scenarios, the initial particle distribution was spread across regions with similar road geometries. This posed a considerable challenge for MapLite in achieving convergence, as it primarily depends on road intersections for localization. Notably, in maps such as `CARLA4' and `Real2' where the road geometry remains consistently similar over extended durations, the MapLite system experiences substantial delays in achieving convergence. In contrast, \coolname{}, which relies on language descriptions of landmarks along the way, has a higher likelihood of encountering a unique set of landmarks, consequently leading to faster global convergence. This observation highlights the significance of our approach, particularly in urban environments with Manhattan's grid-like road layout. Further, the performance of both \coolname \ and \coolname\textit{(L)} is comparable.

\begin{table}[!htp]\centering
\caption{\textbf{Distance(m) to Converge} (lower = better)}\label{tab:convergence}
\scriptsize
\begin{tabular}{lrrrrr}\toprule
\textbf{Scene} &\textbf{CARLA2} &\textbf{CARLA4} &\textbf{Real1} &\textbf{Real2} \\\midrule
\textbf{MapLite} &52.45 &120.45 &46.99 &73.1 \\
\textbf{ALT-Pilot(L)} &29.87 &34.9 &\textbf{24.9} &23.65 \\
\textbf{ALT-Pilot} &\textbf{27.79} &\textbf{32.21} &25.82 &\textbf{23.32} \\
\bottomrule
\end{tabular}
\end{table}

\subsection{Closed Loop Reachability Results}
Table~\ref{tab:closedloop } evaluates performance in closed-loop goal reachability experiments across two CARLA maps. We classify goals as either `near intersection' (within 30m of an intersection) or `not near intersection'. Using five random goals for each category, we measure the distance between the vehicle's final position and the goal.

Capitalizing on language landmark descriptions, \coolname{} achieves superior reachability reachability. For `near intersection' goals, it's on average twice as accurate as MapLite. For `not near intersection' goals, it surpasses MapLite by six times in the `CARLA2' map and ten times in the larger `CARLA1' map.

\begin{table}[!htp]\centering
\caption{\textbf{Closed Loop Reachability.}  Distance(m) between the goal and the final vehicle position.}\label{tab:closedloop }
\scriptsize
\begin{tabular}{lrrrrrr}\toprule
\multicolumn{2}{c}{\multirow{2}{*}{\textbf{Goal Category}}} &\multicolumn{2}{c}{\textbf{CARLA1}} &\multicolumn{2}{c}{\textbf{CARLA2}} \\\cmidrule{3-6}
& &\textbf{MapLite} &\textbf{\coolname} &\textbf{MapLite} &\textbf{\coolname} \\\midrule
\multicolumn{2}{c}{\textbf{ Near Intersection}} &3.22 &\textbf{1.53} &3.896 &\textbf{1.49} \\
\multicolumn{2}{c}{\textbf{Not Near Intersection}} &14.37 &\textbf{1.5675} &11.24 &\textbf{1.76} \\
\bottomrule
\vspace{-5mm}
\end{tabular}
\end{table}

\subsection{Case Studies}

\textbf{False Positive and False Negative Landmarks:}\label{sec:fpandfn}
Table \ref{tab:fpandfn } analyzes our system's resilience to erroneous landmarks, measured by false negatives (FN) and false positives (FP) percentages. For x\% FN, we simulate the omission of x\% of landmark descriptions from OSM, leaving the physical landmarks unchanged. For x\% FP, we modify x\% of landmark descriptions, e.g., changing "signboard" to "fountain". Conducted on the `CARLA1' map (where MapLite had an APE of 10.3 as in Table \ref{tab:ape }), our system remains robust to 40\%-50\% erroneous landmarks. Beyond this threshold, our performance begins to degrade. At around 80\% erroneous landmarks, our system's performance is similar to MapLite, indicating our robustness to gross landmark errors.

\begin{table}[!htp]\centering
\caption{\textbf{False Positive and False Negative Landmarks.} APE and Variance of \coolname \ for various levels of erroneous landmarks compared to MapLite on the same map.}\label{tab:fpandfn }
\scriptsize
\begin{tabular}{lrrrrrr}\toprule
\multicolumn{2}{c}{\textbf{Erroneous Landmarks Type}} &\multicolumn{2}{c}{\textbf{False Negative}} &\multicolumn{2}{c}{\textbf{False Positive}} \\\cmidrule{1-6}
\multicolumn{2}{c}{} &\textbf{APE} &\textbf{Variance} &\textbf{APE} &\textbf{Variance} \\\cmidrule{1-6}
\multicolumn{2}{c}{\textbf{\% of Erroneous Landmarks}} &\textbf{} &\textbf{} &\textbf{} &\textbf{} \\\midrule
\multirow{9}{*}{\textbf{ALT-Pilot}} &\textbf{0} &3.72 &16.51 &3.72 &16.51 \\
&\textbf{10} &3.64 &18.05 &3.42 &17.64 \\
&\textbf{20} &3.8 &24.72 &3.96 &18.96 \\
&\textbf{30} &3.65 &26.01 &3.54 &19.91 \\
&\textbf{40} &4.23 &31.33 &3.7 &20.77 \\
&\textbf{50} &5.89 &47.4 &4.33 &29.32 \\
&\textbf{60} &7.31 &65.22 &5.46 &37.14 \\
&\textbf{70} &8.71 &79.38 &9.11 &55.86 \\
&\textbf{80} &9.58 &97.62 &9.93 &71.68 \\
\midrule
\multirow{2}{*}{\textbf{MapLite}} &\textbf{} &\multicolumn{2}{c}{\textbf{APE}} &\multicolumn{2}{c}{\textbf{Variance}} \\
&\textbf{} &\multicolumn{2}{c}{10.3} &\multicolumn{2}{c}{120.44} \\
\bottomrule
\vspace{-5mm}
\end{tabular}
\end{table}

\begin{figure}[!htb]
    \centering
    \includegraphics[width=0.45\textwidth]{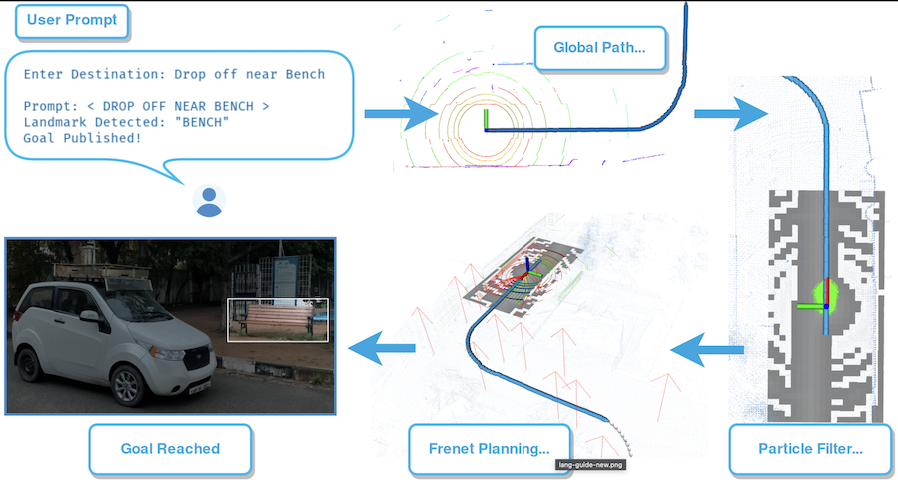}
    \caption{\textbf{Language Guided Navigation in Real World.} Openset language goals can be interpreted and correlated with the language landmarks to find a goal. The \coolname \ system navigates to the goal autonomously.}
    \label{fig:langnav}
    \vspace{-1em}
\end{figure}

\textbf{Real-world Language-goal Navigation:}\label{sec:langnavreal}
Topometric maps augmented with language landmarks introduce a novel application of language-driven navigation. Following a similar approach to~\cite{conceptfusion}, we built a goal selection function, denoted as \textit{goal(`description')}, capable of planning a path to a goal based on an open-ended goal description. An off-the-shelf LLM (GPT-4) is employed to parse natural language goal descriptions into an appropriate function call. For instance, a natural language request such as \textit{``Take me to a place where I can sit"} will be transformed into \textit{goal(``place where I can sit"}). The goal selection function will then identify the closest match among landmarks (possibly a \textit{bench}) corresponding to the description and plan a path to it. Subsequently, the \coolname \ system will autonomously navigate to the specified goal, as demonstrated in Figure \ref{fig:langnav}.

\section{Conclusion}

This paper presents a method for robust navigation and localization using multi-modal maps.
We demonstrated that off-the-shelf vision-language models may be used to construct language-enhanced topometric maps, which can then be consumed by \coolname{} for navigating to arbitrary destinations without the need for HD LiDAR maps.
We implement this on a fullsize real-world autonomous car, demonstrating that our approach is both practical and performant.
Future work may focus on improving the efficiency of the navigation system, and also automating map construction and updates so that all processing is carried out on the vehicle with minimal human intervention.
See our \href{\webpage}{project page} for more details.

\bibliographystyle{IEEEtran}
\bibliography{IEEEtranBST/IEEEabrv, root}

\end{document}